\documentclass[preprint, review, 3p, authoryear, times]{elsarticle}

\makeatletter
\def\ps@pprintTitle{%
 \let\@oddhead\@empty
 \let\@evenhead\@empty
 \let\@oddfoot\@empty
 \let\@evenfoot\@empty
}
 
\usepackage{graphicx}
\usepackage{url}
\usepackage{color, soul}
\usepackage{bm}
\usepackage{amsmath} 
\usepackage{lineno}

\usepackage{geometry}
\usepackage{array}

\usepackage{longtable}
\usepackage{booktabs}
\usepackage{url,lineno,microtype,subcaption, graphicx,float}
\usepackage[table]{xcolor}

\usepackage[utf8]{inputenc}
\usepackage{textcomp} 

\usepackage[]{todonotes}

\begin{document}


\title{
GenAI-powered Multi-Agent Paradigm for Smart Urban Mobility: Opportunities and Challenges for Integrating Large Language Models (LLMs) and Retrieval-Augmented Generation (RAG) with Intelligent Transportation Systems
} 


\tnotetext[t1]{This manuscript has been co-authored by UT-Battelle, LLC, under contract DE-AC05-00OR22725 with the US Department of Energy (DOE). The US government retains and the publisher, by accepting the article for publication, acknowledges that the US government retains a nonexclusive, paid-up, irrevocable, worldwide license to publish or reproduce the published form of this manuscript, or allow others to do so, for US government purposes. DOE will provide public access to these results of federally sponsored research in accordance with the DOE Public Access Plan (http://energy.gov/downloads/doe-public-access-plan).}
\author[CSED]{Haowen Xu} \corref{cor1}\ead{xuh4@ornl.gov}
\author[ESTD]{Jinghui Yuan} \ead{yuanj@ornl.gov}
\author[ESTD]{Anye Zhou} \ead{zhoua@ornl.gov}
\author[ESTD]{Guanhao Xu}\ead{xug1@ornl.gov} 
\author[ESTD]{Wan Li}\ead{liw2@ornl.gov}
\author[UW]{Xuegang (Jeff) Ban}\ead{banx@uw.edu}
\author[TAMU]{Xinyue Ye}\ead{xinyue.ye@tamu.edu}

\cortext[cor1]{Corresponding author.}
\address[CSED]{Computational Urban Sciences Group, Oak Ridge National Laboratory, Oak Ridge, TN 37830, USA}
\address[ESTD]{Buildings and Transportation Science Division, Oak Ridge National Laboratory, Oak Ridge, TN 37830, USA}
 \address[UW]{Department of Civil and Environmental Engineering, University of Washington, Seattle, WA 98195, USA}
 \address[TAMU]{Department of Landscape Architecture and Urban Planning, Texas A\&M University, College Station, TX 77843, USA}
 


\begin{abstract} \label{sec:abstract}
Leveraging recent advances in generative AI, multi-agent systems are increasingly being developed to enhance the functionality and efficiency of smart city applications. This paper explores the transformative potential of large language models (LLMs) and emerging Retrieval-Augmented Generation (RAG) technologies in Intelligent Transportation Systems (ITS), paving the way for innovative solutions to address critical challenges in urban mobility. We begin by providing a comprehensive overview of the current state-of-the-art in mobility data, ITS, and Connected Vehicles (CV) applications. Building on this review, we discuss the rationale behind RAG and examine the opportunities for integrating these Generative AI (GenAI) technologies into the smart mobility sector. We propose a conceptual framework aimed at developing multi-agent systems capable of intelligently and conversationally delivering smart mobility services to urban commuters, transportation operators, and decision-makers. Our approach seeks to foster an autonomous and intelligent approach that (a) promotes science-based advisory to reduce traffic congestion, accidents, and carbon emissions at multiple scales, (b) facilitates public education and engagement in participatory mobility management, and (c) automates specialized transportation management tasks and the development of critical ITS platforms, such as data analytics and interpretation, knowledge representation, and traffic simulations. By integrating LLM and RAG, our approach seeks to overcome the limitations of traditional rule-based multi-agent systems, which rely on fixed knowledge bases and limited reasoning capabilities. This integration paves the way for a more scalable, intuitive, and automated multi-agent paradigm, driving advancements in ITS and urban mobility.

\end{abstract} \label{sec:abstract}
\maketitle


\section{Introduction}
\label{sec:intro}
With rapid urbanization, an estimated 40\% of the population spends at least one hour commuting daily, making urban mobility management critical for smart, sustainable cities \citep{schafer2000future}. Increasing urban populations and automobile numbers have pushed transportation systems to their limits, leading to congestion, accidents, energy waste, and pollution \citep{sperling2009two}. To address these challenges, cities are leveraging advancements in Artificial Intelligence (AI), Internet of Things (IoT), 5G, and computing technologies to transform urban transportation into more efficient, safe, and sustainable systems \citep{sodhro2020towards}. These technological innovations are at the core of developing Intelligent Transportation Systems (ITS), which utilize vast amounts of mobility data from traffic sensors, connected vehicles (CVs), simulations, and crowdsourcing to enable real-time situational awareness and predictive analytics of urban traffic dynamics \citep{li2021emerging,xu2023smart}. These systems optimize infrastructure design, traffic controls, and vehicle operations while providing intelligent advisory services and decision support for travelers and operators\citep{de2016real,mandhare2018intelligent}. Cities worldwide are adopting advanced ITS to create connected transportation environments that maximize safety, mobility, and environmental performance, as seen in initiatives like Japan's VICS and Smartway, Europe's CVIS and COOPERS, and the U.S.'s ITS Strategic Plan \citep{an2011survey}.

Despite the successes of ITS globally, challenges remain in developing scalable, accessible, and interoperable smart mobility solutions \citep{javed2022future}. These challenges hinder the full potential of ITS and CV technologies, especially for diverse urban residents, including commuters, drivers, planners, and policymakers. Smart city technologies must scale to accommodate growing populations and complex systems \citep{bondi2000characteristics}. Modernizing urban transportation requires extensive data processing and analysis, alongside labor-intensive efforts to develop traffic simulations and ITS software components \citep{torre2018big,xu2023smart}. Additionally, the inaccessibility of smart mobility services and transportation data limits effective utilization by city residents \citep{cledou2018taxonomy}. There's a need for a digital assistant capable of helping users discover and access ITS tools and information intuitively. As smart city planning shifts towards human-centric approaches, next-generation ITS and CVs should promote public participation through user-friendly, intuitive technologies that do not require specialized skills \citep{alsayed2024urban,vasilieva2023participatory}. These technologies should enable residents to customize smart mobility services through non-technical interactions.

\begin{figure*}[htb]
 \centering
\includegraphics[width=\textwidth]{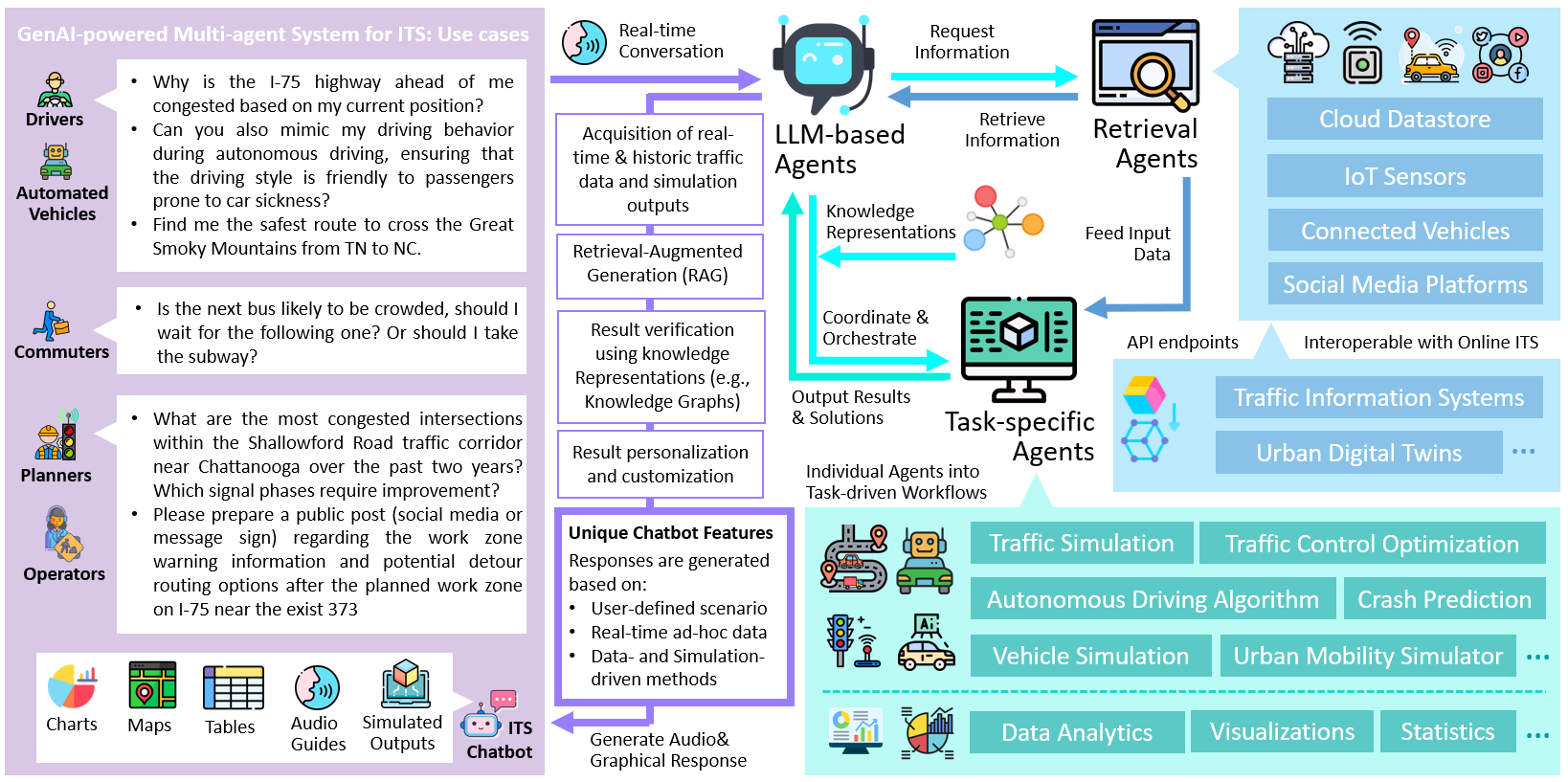}
 \caption{An GenAI-powered Multi-agent System for Intelligent Transportation and Smart Mobility.}
 \label{fig:concepts-design}
\end{figure*}

Building on AI's history in transportation research, there's a growing trend to leverage generative AI (GenAI) technologies to overcome limitations in existing ITS through advanced language understanding, content generation, and reasoning. This paper explores the potential of integrating Large Language Models (LLMs) and Retrieval-Augmented Generation (RAG) technologies to transform ITS and CVs into intelligent multi-agent systems, advancing smart mobility services to be more autonomous, engaging, and human-centric. We review current mobility data and technologies in ITS and CV sectors, and propose a conceptual framework for integrating GenAI with ongoing ITS and CV applications in terms of the following aspects: (a) Promoting science-based advisory for reducing congestion, accidents, and emissions, (b) Enhancing public engagement in mobility management, and (c) Automating transportation management tasks like data analytics and traffic simulations. Our target user groups include drivers of both traditional and automated vehicles, urban commuters who rely on public transit, transportation planners tasked with optimizing infrastructure design and traffic control for improved efficiency and safety, and traffic operators from local government agencies (e.g., state DOT and traffic management center) responsible for emergency response, traffic operations, and asset maintenance. The concept of our proposed framework is illustrated in Figure \ref{fig:concepts-design}.

Compared to conventional chatbots developed using general-purpose LLMs, our proposed LLM-RAG-powered multi-agent system offers the following unique features:
\begin{enumerate}
    \item Responses are generated based on user-defined scenarios that reflect the personalized transportation and mobility needs of specific user groups. This ensures that interactions are directly relevant to the unique context of each user.
    \item Responses are generated using both real-time and historical data (e.g., traffic, weather, infrastructure, public transit, and the user's GPS position), ensuring they reflect the current and situational realities of urban mobility. This enhances decision-making by ensuring the information is timely and context-aware.
    \item Responses are personalized and tailored to the user's specific requirements and preferences, utilizing advanced methodologies, such as data-driven and simulation-based approaches. This ensures that the answers are not generic or Wikipedia-like explanations, but instead are grounded in robust, context-specific analysis of transportation and mobility concepts.
\end{enumerate}
Ultimately, we propose GenAI-powered multi-agent systems that merge ITS developments with emerging LLMs and RAG technologies, enabling more intuitive, efficient, and interoperable solutions for advancing urban mobility management.

\section{Literature Review}
\label{Literature Review}
Over the past decades, Intelligent Transportation Systems (ITS) have evolved alongside advances in information and communications technology (ICT), transforming urban transportation into optimized, digitally enhanced smart mobility. The benefits of ITS and connected vehicle (CV) applications include reducing congestion, enhancing safety, identifying bottlenecks, and optimizing transportation planning and infrastructure design. Previous smart mobility research has focused on integrating AI, advanced simulation, and communication technologies to create more efficient, safer, and sustainable urban transportation systems. This paper reviews and discusses the following key areas:
\begin{description}
\item[Traffic State Forecasting and Vehicle Behavior Prediction] use algorithms and data analytics to predict traffic conditions and vehicle movements, enhancing management, safety, and efficiency.
\item[Advanced Driver Assistance and Intelligent Speed Adaptation] leverage real-time data and AI to guide drivers and adjust vehicle behavior, improving safety and fuel-efficiency.
\item[Traffic Safety and Crash Risk Modeling] employ predictive analytics to assess crash risk, enabling proactive safety measures.
\item[Network Traffic Control and Optimization] manage and adjust traffic flows in real-time to reduce congestion and improve efficiency at system level.
\end{description}

In the following subsections, we will explore the diverse transportation data generated by modern intelligent transportation technologies, which serve as the foundation for advanced traffic analytics and optimization—key to achieving smart mobility systems. We will also examine current research efforts in each application area, providing detailed discussions. This review lays the groundwork for our vision of developing GenAI-powered multi-agent systems for next-generation urban smart mobility. This paper primarily focuses on the planning, management, and optimization of road transportation and urban mobility systems. 
\subsection{Emerging Mobility Data}
\label{subsec:Data}
Mobility data encompasses a wide array of types derived from various technological sources, including IoT-connected transportation infrastructure, connected vehicles, and social media platforms \citep{zhang2011data, singh2022state}. Each of these technologies offers unique insights into different aspects of urban mobility, forming the backbone of advanced data-driven and simulation-driven approaches to sustainable urban transportation planning and mobility management \citep{li2021emerging}.

\subsubsection{Road-side Infrastructure}
The Internet of Things (IoT) has transformed transportation infrastructure by enabling the collection and analysis of vast mobility data, crucial for enhancing traffic management, road safety, public transportation, and reducing environmental impacts \citep{guerrero2018sensor}. IoT integration in roadside infrastructure generates diverse mobility data, each captured by specialized sensors designed to monitor various aspects of vehicular and environmental dynamics \citep{klein2024roadside}.

\textbf{Traffic flow data} includes vehicle volume, speed, and traffic density across road networks. It is crucial for managing congestion, planning road expansions, predicting traffic patterns, and detecting traffic incidents \citep{leduc2008road, xu2021continuous, moriano2024spatiotemporal}. This data is typically obtained by inductive loop sensors, radar systems, or CCTV cameras with advanced analytics \citep{klein2024roadside, zheng2022intelligent, berres2024traffic}. \textbf{Parking data} informs drivers about available parking spaces and optimizes the use of parking facilities \citep{lin2017survey}. It is usually collected through ultrasonic sensors, magnetometers, or cameras that monitor occupancy in real time \citep{shao2022computer, kianpisheh2012smart, dixit2020smart}. \textbf{Environmental data} includes air quality, temperature, and humidity metrics, essential for assessing transportation systems' environmental impact and adapting to weather conditions \citep{asam2015climate, guerrero2018sensor}. This data is gathered by air quality monitors, weather stations, and road surface sensors across transportation networks \citep{mead2013use}.  
\textbf{ Public transit data} offers real-time insights into bus, train, and other transit operations \citep{welch2019big}. It includes vehicle location, passenger numbers, and schedule adherence, enabling transit agencies to optimize services and provide accurate passenger information \citep{koutsopoulos2019transit}.
 
Advancements in cloud computing, digital twins, and cyberinfrastructure have made diverse mobility data more accessible via various wireless communication protocols \citep{ xu2022interactive}. Web service endpoints, along with well-documented metadata, offer new opportunities for AI technologies. These technologies, through machine-to-machine communication, enable automatedv decision-making, data analytics, information sharing, and enhanced visualization capabilities \citep{xu2023smart, xu2024semi,  berres2021multiscale, berres2021explorative}. Since positioning and movement data for many private vehicles and public transit systems are captured through vehicular sensors and onboard units rather than roadside infrastructure, these aspects will be discussed in the next subsection.




\subsubsection{Connected Vehicles}
A CV is an automobile equipped with internet access and wireless communication capabilities, allowing it to send and receive data, interact with other vehicles, smart infrastructure, and cloud services \citep{guerrero2015integration}. This connectivity enhances safety, efficiency, and convenience by enabling real-time information exchange, supporting advanced driver assistance systems (ADAS), facilitating vehicle-to-vehicle (V2V) and vehicle-to-infrastructure (V2I) communication, and enabling over-the-air updates \citep{siegel2017survey}. Both vehicle manufacturers and transportation agencies emphasize that CV technology can significantly reduce fatalities and serious injuries from vehicle crashes while promoting eco-driving to reduce fuel consumption and carbon emissions \citep{gruyer2021connected}. Public transit data, collected through Automatic Vehicle Location (AVL) systems, passenger counting systems, and Transit Signal Priority (TSP) systems, further supports transportation management by tracking vehicle positions and monitoring passenger numbers \citep{hellinga2011estimating, darsena2023enabling}.

As a foundational element of ITS and the broader IoT ecosystem, CVs generate vast amounts of valuable traffic data through advanced communication technology and vehicular sensors \citep{siegel2017survey}. GPS loggers and cameras are key components, enabling data exchanges across V2V and V2I networks and facilitating the transmission of basic safety messages (BSM) between vehicles and roadside equipment (RSE). This data includes vehicle trajectories (e.g., speed, location, direction), driver-vehicle interactions (e.g., steering, pedal usage, gear shifts), and contextual factors (e.g., weather) \citep{parikh2017vehicle, harding2014vehicle, siegel2017survey}. More advanced data from vehicular radar, LiDAR, ultrasonic sensors, and inertial measurement units (IMU) enables high-resolution 3D mapping, advanced parking support, and precise measurement of vehicle dynamics \citep{nidamanuri2021progressive}. Several studies have comprehensively reviewed CV data, its sources, and applications in vehicle technology and urban mobility \citep{siegel2017survey, li2021emerging}.  

In the rapidly evolving landscape of ITS and smart mobility, companies like Wejo, INRIX, Otonomo, Mobileye, HERE Technologies, TomTom, StreetLight Data, CARMERA, DeepMap, Teralytics, and Arity have become key players in collecting, analyzing, and distributing connected vehicle data. These companies provide extensive datasets critical for advancing urban transportation solutions and research. Leveraging data from vehicle telemetry, GPS, cameras, LiDAR, radar, and mobile networks, they offer insights into traffic flow, vehicle trajectories, driving behavior, and road conditions. This data drives the development of advanced traffic management systems, autonomous driving technologies, and smart city infrastructure, enabling more efficient, safe, and sustainable urban mobility. Practical applications of CV data include systems like the Intersection Collision Warning System (ICWS), which alerts drivers to potential crash risks at intersections \citep{usdot_highways}, and the Green Light Optimal Speed Advisory (GLOSA), which advises vehicles on optimal speeds to reduce fuel consumption, carbon emissions, and traffic congestion \citep{suzuki2018new, xu2023mobile}. In addition, both real-time and historical CV data can be accessed via API endpoints and real-time data feeds. This opens pathways for AI agents to utilize this information to enhance smart city initiatives.

\subsubsection{Social Media Data}
In recent decades, social media platforms have become valuable sources of mobility data, offering real-time insights into human movement, transportation preferences, and traffic conditions \citep{zheng2015big}. Geo-referenced social media plays a crucial role in ITS by providing dynamic information that improves traffic management and operation, enhances road safety, and optimizes public transit \citep{torbaghan2022understanding}. The mobility data from social media can be categorized as follows:

\textbf{Geotagged posts} are social media posts with geographic coordinates, offering precise snapshots of user locations and movements. Platforms like Twitter generate vast geospatial data, valuable for understanding popular destinations, tracing movement patterns, and analyzing urban flow \citep{torbaghan2022understanding, zheng2015big}. For ITS, geotagged data helps identify high-traffic areas, optimize routes, and manage crowd dynamics, leading to more efficient traffic management.
\text{Mobility} trends and patterns are uncovered by aggregating and analyzing geo-referenced user-generated content. By examining hashtags, post frequencies, and temporal patterns, researchers can identify urban mobility trends such as peak travel times and preferred routes \citep{martin2019leveraging}. This data enables ITS to predict congestion, adapt transit services, and enhance urban transportation efficiency.
\textbf {Sentiment and opinion data} related to transportation, analyzed from posts, comments, and hashtags, reflects public sentiment about transportation services, infrastructure, and policies \citep{qi2020framework}. Platforms like Twitter, Reddit, and Facebook provide insights into user experiences, highlighting areas for improvement and successful initiatives, which are crucial for aligning ITS with public expectations.
\textbf{Crowdsourced traffic information} from social media is pivotal for ITS. Users report real-time traffic conditions, such as incidents, work zones, and congestions, on platforms like Twitter, while apps like Waze enable direct sharing of traffic updates \citep{adler2014estimate}. This data supports dynamic traffic management, allowing operators to adapt swiftly to changing conditions, enhancing road safety and improving travel efficiency \citep{liu2024evaluating}.
 
In summary, social media platforms produce diverse mobility data that is essential for developing and optimizing intelligent transportation systems. This data, accessible in real-time through various API endpoints, presents an opportunity for an intelligent multi-agent system. Such a system allows AI agents to access critical, up-to-date social media information, enhancing situational awareness and optimizing decision-making in smart mobility services, particularly in traffic safety, route planning, and public transit trip planning.

\subsection{Popular Intelligent Transportation Applications}
\label{sec:ITS_applications}
Driven by the vast and varied traffic data, novel methodologies have been integrated into ITS to enhance urban mobility from multiple perspectives, which are individually reviewed and discussed through the following subsections. With advancements in ICT and computing technologies, the algorithms driving these ITS methodologies can now be accessed via wireless communication protocols and API endpoints through cyber delivery. This development creates opportunities for AI agents to leverage these methods as web or cloud services, enabling automated analytics and optimizations when they are tasked to provide smart mobility services to urban residents. 

\subsubsection{Traffic State Forecasting and Vehicle Behavior Prediction}
\label{subsec:accident}
Traffic state forecasting and behavior prediction are crucial for developing strategies that enhance road safety, reduce congestion, and improve traffic efficiency. Traffic state forecasting predicts future conditions like flow, density, and speed by leveraging historical and real-time data from sources such as loop detectors, cameras, probing devices, and social media \citep{kumar2021applications}. Vehicle behavior prediction focuses on forecasting the trajectories of surrounding vehicles using data from historical trajectories and traffic environments gathered through sensors and communication networks.

Algorithms for traffic state forecasting and vehicle behavior prediction can be broadly categorized into model-based and learning-based methods. Traditional model-based methods include statistical time series models \citep{Antoniou2013DynamicDL, Wang2005RealtimeFT, Xing2020PersonalizedVT, Wang2022ARIMAMA, li2020short}, rule-based models \citep{Szeto2009MultivariateTF, Zhong2013StochasticCT, Zhou2023CarFollowingBO, Woo2016DynamicPF}, and probabilistic models \citep{Wang2014NewBC, Wang2021ABI, Zhang2020ResearchOT, lin2017road}. While these methods can perform well in specific scenarios, their rigid structures and underlying assumptions can limit effectiveness, especially when faced with unmodeled uncertainties.

In contrast, advancements in computing power and data availability have driven the rise of model-free, learning-based methods. These methods, often based on deep neural networks (DNNs), offer more accurate and robust performance under uncertain conditions. DNNs used in traffic forecasting and behavior prediction can handle multi-modal inputs, generate high-dimensional outputs, and learn spatial dependencies (e.g., convolutional neural networks) \citep{Lin2022VehicleTP, Do2018SurveyON, KhajehHosseini2019TrafficPU,subramaniyan2022hybrid,li2020real,ke2020two}, capture social interactions between agents (e.g., graph neural networks) \citep{Lin2017PredictingSH, Cui2020LearningTA, Mo2021GraphAR, Schmidt2022CRATPredVT}, and model time-series dependencies (e.g., recurrent neural networks, attention mechanisms) \citep{Cui2020StackedBA, Xu2022LeveragingTM, Liu2021ProactiveLC, Zhou2023CooperativeCO}. Incorporating physical principles into these learning-based methods further enhances DNN generalizability, leading to more reliable applications \citep{Mo2020APD, Geng2023APT, Huang2020PhysicsID}.

The emergence of LLMs presents an opportunity to further improve forecasting and prediction by integrating expert feedback from traffic professionals. This feedback can refine DNN training and offer meaningful interpretations of input features, enhancing the trustworthiness and interpretability of the overall workflow.

\subsubsection{Driving Advisory and Intelligent Speed Adaptation}
\label{subsec:CV}
Advanced Driving Assistance Systems (ADAS) are sophisticated technologies designed to enhance driver decision-making and road safety by providing real-time guidance \citep{ameta2023advanced}. Utilizing data from GPS, traffic monitoring systems, and vehicle sensors, ADAS offers advice on lane usage, merging, speed adjustments, and route selection, reducing driver workload and improving safety \citep{antony2021advanced}.

Intelligent Speed Adaptation (ISA) helps drivers maintain appropriate speeds by either advising them of speed limits or automatically adjusting the vehicle's speed to comply with legal limits, enhancing safety and fuel efficiency \citep{ryan2019intelligent, newsome2024potential}. ISA systems use GPS, digital maps, and onboard cameras to determine speed limits, functioning either as advisory systems or as automatic speed controllers \citep{archer2001estimating, carsten2006intelligent}. The goal of ISA is to reduce speed-related traffic accidents and improve road safety.

Building on vehicle communication technologies, ADAS and ISA have been widely adopted to optimize Connected Automated Vehicles (CAVs) \citep{nidamanuri2021progressive}. CAV control varies by levels of cooperation and automation. In connected but non-automated scenarios, advisory systems compute energy-saving or congestion-mitigating trajectories for human drivers \citep{Wan2016OptimalSA, Xu2024AME, Suramardhana2014ADG}. In automated but non-connected scenarios, ADAS ensures smooth vehicle operations using onboard sensors \citep{Zhou2022SignificanceOL,Zhou2022IncorporatingDR,Zhou2022CongestionmitigatingMD,Zhou2023ModelPC}. Fully connected and automated systems exhibit four classes of intelligent vehicle control based on the Society of Automobile Engineers standards: (i) status sharing (class A), where CAVs share trajectory information to improve traffic flow \citep{Wang2018CooperativeAC, Zhou2023CooperativeCO, Zhou2024ImplicationsOS, Zhou2022RobustCS}; (ii) intent sharing (class B), where future plans and control decisions are shared to enhance anticipation and situational awareness \citep{Wang2023EnergyCentricCO, Wang2023CooperativeMS, Liu2023ProactiveLC, Liu2021ProactiveLC}; (iii) agreement seeking (class C), where multiple vehicles negotiate cooperative plans to balance interests \citep{Hyeon2024ALA,Hyeon2023PotentialES,Hyeon2023DecisionMakingSU}; and (iv) prescriptive (class D), where CAVs follow a centralized plan to achieve system-level optimization \citep{Wang2024RobustCC, Zhou2022CooperativeSI, Wang2022RealtimeDA}.

Despite their effectiveness, challenges remain in personalizing autonomous driving systems, interpreting AI decisions, and fostering trustworthy human-machine interactions. Multi-modal LLMs offer potential solutions by integrating human input and articulating AI and human rationales in a bidirectional manner.

\subsubsection{Traffic Safety and Crash Risk Modeling and Prediction}
\label{subsec:accident}
Traffic safety applications involve the use of statistical and machine learning techniques to assess and predict the frequency and likelihood of traffic crashes \citep{chand2021road}. Traffic safety models are initially developed to identify patterns and factors contributing to crashes, such as driver behavior, road conditions, weather, and traffic flow characteristics, with the goal of enhancing road safety \citep{theofilatos2014review}. Later on, by incorporating historical crash data and real-time information, real-time crash risk prediction models were developed predict potential high-risk situations and locations. In addition, traffic conflict based safety models have been popular to address the limitation of rareness and randomness of crash data. In general, there are three types of traffic safety models that have been widely developed and implementated in real world.

Crash frequency prediction models, also known as safety performance functions (SPF), estimate the expected number of crashes at specific locations over a given period, enabling transportation planners to identify high-risk areas and allocate resources effectively. These models often incorporate various factors such as aggregated traffic volume, road geometry, and traffic control characteristics to achieve better prediction accuracy \citep{khattak2024analysis}.

Crash severity models focus on predicting the potential outcomes of crashes, i.e., the severity of crashes. These models typically assess the impact of factors such as vehicle speed, driver and occupant characteristics, collision type, and vehicle safety features \citep{qin2013analysis}. Understanding crash severity helps in the development of targeted safety interventions (e.g., driver education and advanced vehicle safety features) and infrastructure improvements that can mitigate the most serious consequences of traffic accidents \citep{khattak2020bayesian}.
    
Real-time crash risk evaluation and prediction models aim to model the relationship between crash likelihood and the corresponding real-time traffic flow characteristics, and then predict the probability of crash occurrence in real-time. The prediction results can not only provide dynamic warnings or interventions designed to prevent potential crashes for drivers but also provide better predictive insights for traffic operators to improve the emergency response and reduce the incident clearance time. \citep{yuan2020modeling, yuan2018approach, yu2020incorporating, li2020real, cheng2022crash, yue2020augmentation}.

Surrogate safety models provide a new perspective for traffic safety modeling and prediction given traffic crashes are extremely rare events, which is more scalable and implementable. Modeling the relationship between various surrogate safety measures (e.g., time-to-collision, post-encroachment time, deceleration rate to avoid a crash) and real-time driver and traffic flow characteristics can effectively predict the potential near-crashes before it happens. \citep{abdel2022using, xing2020time, wang2021review}.

These applications are widely used in urban planning, traffic management systems, advanced driver-assistance systems (ADAS) and autonomous driving system, contributing to safer road networks and reduced accident rates.

\subsubsection{Network Traffic Control and Optimization}
\label{subsec:control}
While driving advisory systems and intelligent vehicle control enhance mobility at the individual vehicle level, network traffic control strategies are essential for optimizing entire transportation systems. These strategies improve travel times, reduce fuel consumption, and lower emissions, supporting sustainable urban mobility \citep{ugirumurera2020high, xu2021visualizing,wang2022network, li2022network}. This process involves the systematic management and regulation of traffic flows across road networks, using real-time data and advanced algorithms to optimize traffic signals, reroute vehicles, and adjust traffic patterns.

Among various optimization strategies, traffic signal control is the most widely implemented. Traditional fixed-time traffic signal control, based on historical data, often fails to adapt to real-time traffic variations, leading to inefficiencies. Adaptive traffic signal control systems address these limitations by using real-time data from sensors and cameras to dynamically adjust signal timings, optimizing flow and reducing delays \citep{wang2018review,abdulhai2003reinforcement, li2018connected, li2022point}. Traffic signal coordination synchronizes signals along major corridors, minimizing stops and maintaining continuous traffic flow \citep{bazzan2005distributed,putha2012comparing, li2020connected}. Transit signal priority (TSP) systems further enhance efficiency by adjusting signals to reduce delays for public transit vehicles \citep{smith2005transit,he2014multi,yu2020incorporating}.

Ramp metering is another key strategy, regulating vehicle entry onto freeways to maintain optimal flow and prevent congestion \citep{papageorgiou1991alinea,papageorgiou2002freeway}. It has been shown to enhance freeway throughput, increase vehicle speeds, and improve safety by reducing stop-and-go conditions \citep{haj1995ramp,lee2006quantifying,smaragdis2004flow}. Effective ramp metering relies on real-time data, predictive modeling, and adaptive algorithms \citep{ma2020statistical}.

The Macroscopic Fundamental Diagram (MFD), which gives the relationship between flow, density,and speed at a network level\citep{xu2020analytical,xu2024impact}, has emerged as a valuable tool for managing congestion across entire networks. MFD-based strategies, such as perimeter flow control, congestion pricing, urban road space allocation, regional vehicle routing, and area-wide signal control, are effective in optimizing network-wide traffic flow \citep{aboudolas2013perimeter,zhou2021model,simoni2015marginal,xu2023non,xu2023opposing,yildirimoglu2015equilibrium,yan2016extended}. Implementing these strategies requires robust data collection and advanced analytics to interpret MFD data effectively.

Other network traffic control methods include Variable Speed Limits (VSL), which adjust speed limits based on real-time conditions \citep{lin2004exploring}, and Integrated Corridor Management (ICM), which optimizes overall corridor performance by managing all available infrastructure and resources efficiently \citep{hashemi2016real}. These strategies, often dependent on extensive data collection and analysis, can be further enhanced by integrating LLMs.

\section{
The Raise of Multi-agent Systems
}\label{sec:mas}
A multi-agent system is a framework where multiple autonomous agents, each with specific capabilities, work together through coordination to achieve a common goal or solve complex problems \citep{vogel2020multi}. These agents interact, communicate, and emulate collaborative behaviors found in human and natural systems. Enhanced by emerging AI technologies and knowledge engineering, they now undertake increasingly specialized tasks \citep{khosla2012engineering}. Multi-agent systems have gained traction in research and applications requiring sophisticated problem-solving, such as smart manufacturing and healthcare \citep{yahouni2021smart, shakshuki2015multi}.
The growing complexity of challenges in these domains has driven the evolution of multi-agent systems alongside Generative AI (GenAI) technologies, including LLMs and RAG \citep{de2023emergent}. By leveraging the strengths of specialized agents working together, AI-powered multi-agent systems can more effectively and interactively address complex problems that require deep domain knowledge.

A typical GenAI-powered multi-agent system includes several key components: (a) LLM agents that excel in natural language understanding and generation, (b) retrieval agents that fetch relevant data from external sources, and (c) task-specific agents that handle specialized functions like sentiment analysis, optimization, or decision-making. Together, these agents, guided by domain knowledge (e.g., ontologies and knowledge graphs), form a cohesive unit capable of addressing intricate problems more effectively than a single model alone.



\label{sec:llm}
\subsection{LLM Agents: Origin and Capabilities}
\label{subsec:transformers}
In recent years, pre-trained Large Language Models (LLMs) have significantly expanded AI's capabilities in natural language understanding, reasoning, and generation \citep{yang2024harnessing}. Generative Pre-trained Transformer (GPT) models, developed using the Transformer architecture, have become increasingly popular in both industry and academia \citep{raiaan2024review}. Notable examples include OpenAI's ChatGPT, GPT-3, BERT, T5, and RoBERTa. These models excel in various applications, from conversational agents to content generation, due to their ability to leverage vast textual data to capture nuanced linguistic patterns and contextual relationships. This sets them apart from traditional deep learning methods, which often require task-specific training and struggle with understanding broader contexts.

The flexibility and robustness of LLMs have led to their growing use in urban science research. These models are being used to analyze large volumes of textual data, extract meaningful insights, and support decision-making across sectors like building management \citep{jiang2024eplus, rane2023integrating}, natural hazard mitigation \citep{chandra2024decision,samuel2024application}, energy system management \citep{Choi2024, lai2024bert4st}, and freight system optimization \citep{tupayachi2024towards}. LLMs' ability to process and interpret complex datasets and technical documents with minimal supervision is proving invaluable for addressing big data challenges in these fields.

In an GenAI-powered multi-agent system, a LLM agent typically serves as a specialized entity that processes natural language data and generates new content. The role of an LLM-agent can vary depending on the specific application. In the context of augmenting the existing ITS technologies, LLM agents can facilitate the following functions:

\begin{itemize}   
    \item \textbf{Natural Language Understanding:} The LLM-agent interprets and understands human language inputs, allowing it to communicate effectively with both human users and other agents in the system.
    \item \textbf{Task Automation:} The LLM-agent can automate tasks that involve language processing, such as drafting reports, generating responses, or summarizing information, thereby reducing the workload on human agents or other system components.
    \item \textbf{Mediation and Coordination:} In a multi-agent system, the LLM-agent can facilitate communication and coordination between different agents by translating, summarizing, or contextualizing information exchanged within the system.
    \item \textbf{User Interaction:} The LLM-agent often serves as the primary interface for human users, enabling intuitive and conversational interactions, which enhances user experience and accessibility in complex systems.    
\end{itemize}
Overall, the LLM-agent plays a crucial role in enhancing the cognitive capabilities of a multi-agent system, enabling more sophisticated, context-aware, and human-centric interactions.


\subsection{Retrieval Agent for RAG}
\label{subsec:AI-based}

Retrieval agents was proposed to address the limitations of the early versions of LLM agents in dealing with domain-specific queries requiring specialized knowledge \citep{hu2023survey}. Since these general-purpose models, trained on static text, often struggle to answer domain-specific questions or access the most up-to-date information, making them vulnerable to hallucinations \citep{zaki2024mascqa}. Unlike the computationally expensive fine-tuning processes, which often requires a significant amounts of data to tune the parameter of a LLM, the recent emerging RAG technique presents a cost-effective solution to enhance the language models' knowledge base by enabling an external information retrieval system. The technique expand the general-purpose LLMs' ability to provide accurate and contextually relevant responses by leveraging external knowledge sources that are specialized and domain-specific. 

RAG works by first retrieving pertinent documents from a vector database, which indexes and searches data based on vector embedding, and then generating a response using both the retrieved information and the model's internal knowledge \citep{cakaloglu2020text}. The vector database concept is crucial as it allows efficient and effective retrieval of semantically similar data points, thereby ensuring that the most relevant information is utilized during response generation. Studies have shown that integrating retrieval mechanisms with generative models can significantly improve factual accuracy and reduce the incidence of hallucinations in model outputs \citep{lewis2020retrieval,karpukhin2020dense}.

During this process, the retrieval agent is crucial for sourcing information. It locates and gathers knowledge and data from various repositories, both structured and unstructured. The agent then filters and organizes the data based on relevance, context, or specific criteria, ensuring that only the most pertinent and accurate information is presented or used. The major components of a retrieval agent typically include the (a) an interface layer for communication with other agents or users, (b) a query processor that interprets and formulates queries, and (c) a data retrieval engine that interacts with databases or other storage systems to fetch the required information.    

To support these functions, different types of database technologies can be leveraged based on the nature of the data and retrieval requirements. Among different database technologies, \textbf{Relational databases} are often used for structured data retrieval, enabling efficient querying and management of tabular data with well-defined relationships. \textbf {Graph databases} excel in scenarios where the relationships between data points are complex and require traversal of nodes, such as in topological data, such as a social network or road network. \textbf{Vector databases} are particularly effective for handling unstructured or high-dimensional data, such as embeddings from natural language processing models, enabling similarity search and nearest-neighbor retrieval. By integrating these database technologies, a retrieval agent can be tailored to handle diverse data types and retrieval tasks, ensuring robust and efficient information retrieval in a multi-agent system. 

Recently, several frameworks, including LangChain, Hugging Face, and LlamaIndex (formerly GPT Index), have emerged to provide robust pipelines that seamlessly connect LLM-based agents with various external data sources, particularly databases and APIs \citep{topsakal2023creating, shen2024hugginggpt, pinheiro2023construction}. These frameworks enable the development of powerful retrieval agents that can effectively interact with and process data from diverse sources.

\subsection{Task-Specific Agents}
\label{subsec:AI-based}
In a multi-agent system, task-specific agents are specialized entities designed to perform distinct and specialized roles that are tailored to execute particular tasks. These agents are empowered by specific capabilities, knowledge, and behaviors that align with their assigned roles. They operate autonomously or in coordination with other agents to achieve localized goals that contribute to the overall objectives of the multi-agent system and its target workflow.

In augmenting ITS and smart urban mobility, task-specific agents integrate specialized transportation tools and methodologies, such as agent-based traffic simulations and predictive models for traffic flow, accidents, and driver behavior. These agents can be developed using core ITS components, which are reviewed as examples for the intelligent transportation and smart mobility applications in Section \ref{sec:ITS_applications}. With advancements in cloud and edge computing, features like traffic simulations, vehicle simulators, and predictive analytics can be hosted as web services and delivered via wireless networks. These services enable the creation of task-specific agents using computing paradigms like Service-Oriented Architecture (SOA) and Microservices Architecture, facilitating seamless and interoperable integration with other agents in a multi-agent framework.

\section{Multi-agent Framework for ITS} \label{sec:oppoertunity}
This section reviews the current advancements of GenAI-powered multi-agent systems in research, highlighting their transformative potential. We then explore opportunities enabled by integrating LLM and RAG into multi-agent systems to address the challenges and limitations of traditional ITS applications, as discussed in Section \ref{sec:intro}.

\subsection{Multi-agent Systems for Operations Research}
As LLMs and RAG have been increasingly used in urban research in recent years, a few studies have developed GenAI-powered multi-agent systems to facilitate urban management \citep{xu2024leveraging}, especially in the energy systems and smart grid sector. However, their applications in intelligent transportation and smart mobility remains rare.

\citet{gamage2024multi} introduce a multi-agent chatbot architecture within the La Trobe Energy AI Platform (LEAP) to enhance decision-making in net-zero emissions energy systems. This architecture integrates LLMs and RAG to improve data processing and retrieval, featuring four specialized agents—Observer, Knowledge Retriever, Behavior Analyzer, and Visualizer—responsible for anomaly detection, data visualization, behavioral analysis, and information retrieval. By leveraging RAG, the chatbot delivers contextually relevant responses, optimizing decision support in dynamic energy environments aimed at net-zero carbon emissions.
\citet{choi2024egridgpt} present eGridGPT, a generative AI-powered platform by the National Renewable Energy Laboratory (NREL) to support decision-making in power grid control rooms. Using LLMs and RAG, eGridGPT enhances situational awareness and decision-making by analyzing real-time data, simulating scenarios, and generating actionable recommendations. The system integrates digital twins for simulation-based validation, ensuring accurate and reliable outputs. eGridGPT's multi-agent system streamlines operations, reduces decision-making time, and improves complex energy system management, making it crucial for future grid operations balancing renewable energy integration with grid stability.

\subsection{LLM and RAG in Multi-agent System: Opportunities for ITS}
Traditional multi-agent systems (MAS) were designed to consist of autonomous agents that performed specific tasks through rule-based reasoning and predefined communication protocols \citep{ferber1999multi}. They played key roles in distributed problem-solving across various domains like traffic management and robotics. However, these systems were limited by their fixed knowledge bases, lack of advanced learning capabilities, and inability to process or generate natural language, making them less adaptable and context-aware \citep{wooldridge2009introduction, shoham2008multiagent}. The recent integration of LLMs and RAG has significantly expanded MAS capabilities by enabling advanced capabilities for natural language processing, reasoning, and dynamic knowledge retrieval, thus making these systems more versatile and autonomous in complex, real-world scenarios \citep{gurcan2024llm}.

\subsubsection{Enhancing System Scalability}
In a GenAI-powered multi-agent system, LLM-based agents, equipped with advanced language reasoning and dynamic knowledge capabilities, can autonomously manage repetitive, time-consuming, and labor-intensive tasks essential for developing and maintaining an ITS. These agents are particularly effective in handling the complexities of expansive urban areas with diverse functions, ensuring the system's scalability and efficiency.
\begin{description}
\item[Autonomous Traffic Data Analytic and Interpretation:] Building on previous studies that explore LLM's capabilities for automating data analytics and interpretations \citep{li2023autonomous, zheng2023chatgpt}, we propose that LLM-based agents, integrated with a knowledge graph and RAG, can autonomously process and interpret vast amounts of traffic data. These agents can analyze both real-time and historical traffic data, identify patterns, and generate insights to optimize traffic flow, reduce congestion, and enhance safety. The domain-specific capability of LLMs to extract and interpret patterns in diverse traffic data can be enabled through the incorporation of a knowledge graph and ontological learning \citep{babaei2023llms4ol}. By automating these analytical tasks, the system can continuously adapt to changing traffic conditions without requiring constant human intervention and supervision.
\item[Automated Software and Simulation Developments:] 
As previous studies have explored LLMs' potential for completing complex software engineering tasks \citep{hou2023large, ozkaya2023application}, these AI-powered agents can also automate the development of essential ITS components, including data stores, software modules, and traffic simulation scenarios \citep{xu2024leveraging}. By automatically generating and refining these components based on data and requirements, the system can rapidly deploy ITS solutions across large urban areas. This automation not only accelerates the deployment process but also ensures that the ITS is continually updated with the latest advancements in traffic management technologies.
\end{description}


\subsubsection{Promoting Smart Mobility Service Accessibility}
In addition to their exceptional reasoning and content generation capabilities, LLM-based agents enhanced with RAG can revolutionize traditional ITS by greatly improving user interaction and accessibility. With their advanced natural language understanding, these agents can engage users through natural, human-like conversations, eliminating the need for technical expertise in programming or transportation. By interpreting users' mobility needs, such as reducing commute times, alleviating traffic congestion, or enhancing safety, LLM-based agents can map these requirements to the appropriate task-specific agents, such as those responsible for traffic simulations, flow optimizations, or crash predictions.
\begin{description}
\item[Intelligent Task Coordination:] 
In a multi-agent system, these agents serve as orchestrators, seamlessly integrating and coordinating the activities of various task-specific agents to complete complex decision support tasks. For example, when a user requests to optimize their daily commute, the LLM-based agent interprets the need and delegates tasks to agents specializing in traffic data analysis, route optimization, and traffic simulation. The result is a well-coordinated response that leverages the expertise of multiple agents to deliver a comprehensive, intelligent solution for smart mobility.
\item[ChatBot for Intelligent Driver Assistance:] 
Through conversational interaction, drivers can use these agents to receive guidance on optimal routes, warnings about potential hazards, and advice on improving fuel efficiency. Recent advancements in ontological learning and RAG allow these agents to handle more sophisticated queries by leveraging task-specific agents and real-time traffic data to simulate or identify high-congestion areas and safety risks. This conversational interface enhances the driving experience and makes ITS features accessible to a broader audience, including those unfamiliar with advanced technology.
\end{description}

\subsubsection{Improving Human-centric Mobility Management}
In a generative AI-powered multi-agent system, LLM-based agents enhanced with RAG can transform Human-centric Mobility Management within ITS. These agents, with their advanced language understanding, can effectively interpret diverse user needs, including those of urban commuters, drivers, and transportation planners. By processing everyday language or casual queries, they extract valuable insights reflecting users' concerns, preferences, and expectations.

Leveraging their language capabilities, LLM-based agents can automatically identify relevant data sources and knowledge bases from various online resources. They can generate precise queries using text-to-query conversions, or employ self-query retrievers to deliver accurate traffic data, insights, and explanations for specialized urban mobility and transportation management tasks.
\begin{description}
 \item[Facilitating Crowd-sensing and Crowd-sourcing through Conversations:] LLM-based agents are essential for facilitating crowd-sensing and crowd-sourcing of vital information through conversational interactions with chatbots. These agents can gather real-time feedback, complaints, and suggestions from urban residents about the transportation system, helping ITS stay responsive to the public's evolving needs. This real-time feedback loop enables transportation planners to make data-driven decisions that align with the community's actual experiences and concerns.
 \item[Expert System for Public Education and Engagement:] 
By querying versatile traffic data and knowledge bases hosted through public available web services, these agents can deliver clear, easy-to-understand explanations to urban residents. This empowers the public with knowledge about traffic patterns, safety measures, and mobility options, fostering a more informed and engaged community. The accessibility of these expert systems ensures that even those without technical backgrounds can grasp complex transportation issues, promoting greater public participation in urban mobility planning.
\end{description}


\section{Conceptual Framework: Design and Challenges}
\label{sec:CFTA}
\begin{figure*}[htb]
 \centering
\includegraphics[width=\textwidth]{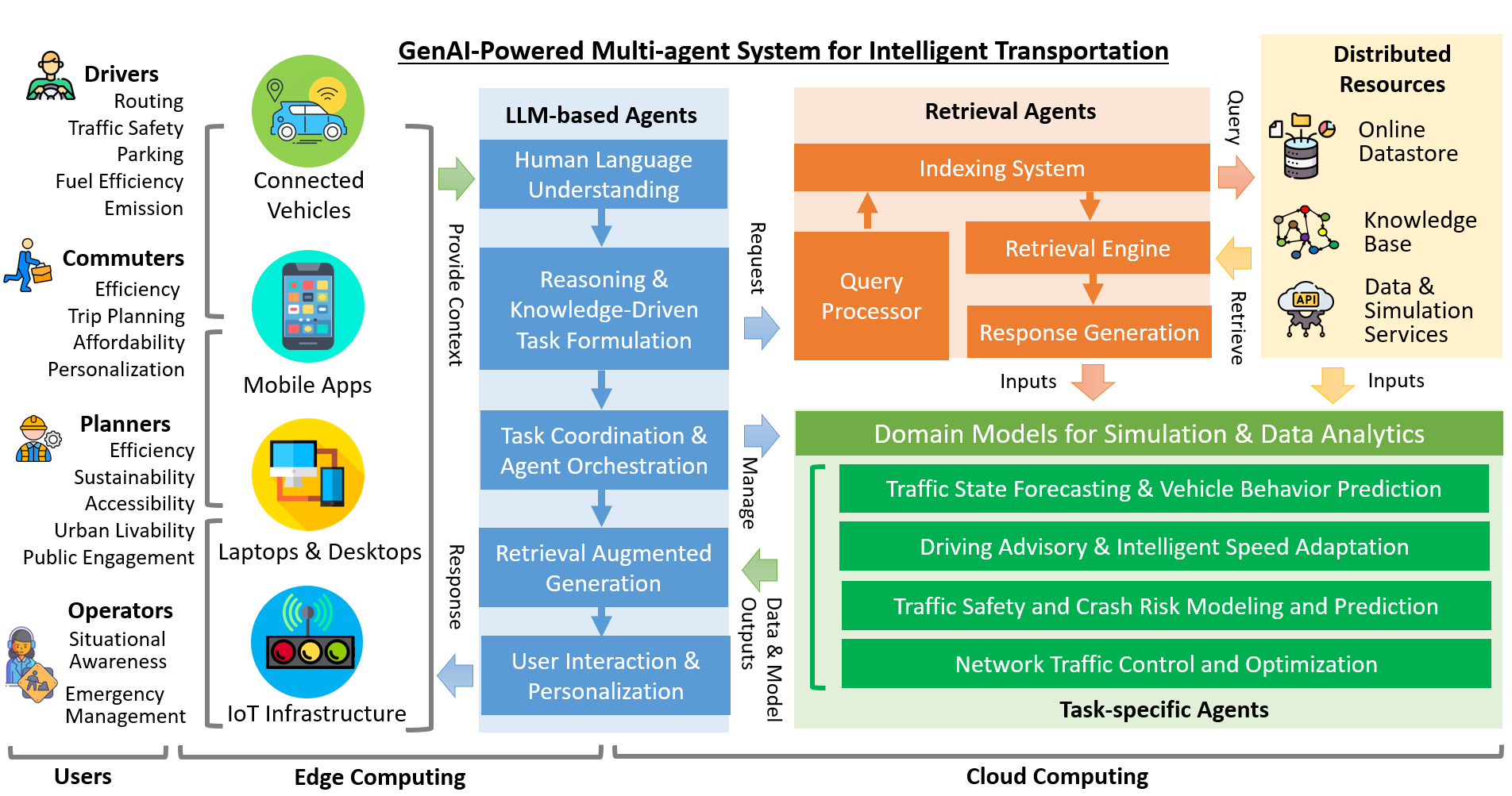}
 \caption{Overall system design to fulfil the needs for various system users in the urban mobility systems. The system is designed to fulfil the needs for 4 types of targeted users, including drivers (in both traditional and automated vehicles), commuters of public transit, transportation planners, and traffic operators from the government bodies.}
 \label{fig:overall-design}
\end{figure*}

We propose a conceptual framework for designing a GenAI-powered Multi-agent System to enhance smart urban mobility (see Figure \ref{fig:overall-design}
), with its components detailed in Section \ref{sec:mas}. 
\subsection{Overall System Design}
Conceptually, in this framework, LLM-based agents are central to user interaction, reasoning, and task formulation using knowledge representations like ontologies and knowledge graphs. They coordinate tasks, orchestrate agents, and deliver personalized, science-based results through RAG. Retrieval agents work with LLM-based agents to fetch essential transportation data from distributed resources, including government portals, third-party data stores, and transportation digital twins. Task-specific agents, equipped with domain models for simulations and analytics, then generate scientific outputs to meet user requests, such as optimizing routes to reduce carbon emissions, finding efficient multi-modal transportation, and optimizing IoT-based traffic controls.

This flexible framework is adaptable across various software development environments and industry-standard stacks, addressing intelligent transportation needs with a plug-and-play approach that connects LLM-based agents to task-specific agents for simulation and data analytics. Leveraging modern DevOps technologies, these modular agents can be deployed and scaled across edge and cloud architectures. LLM-based agents can be implemented using APIs like ChatGPT, Hugging Face Transformers, and Google's Bard, enabling chatbots on both server-side and client-side platforms. While external knowledge sources often eliminate the need for fine-tuning, it may still be preferred for specific outputs or user interactions. Incorporating \textbf{Transportation Knowledge Representations}, such as ontologies and knowledge graphs, guides LLMs in task sequencing and querying the correct datastores and APIs. Retrieval agents, built with frameworks like LangChain and LlamaIndex, connect LLMs to data sources, while task-specific agents host simulations and analytical models as web services, documented in the knowledge representations for seamless querying and access.

\subsection{Potential Challenges}
In this paper, we highlight three methodological challenges associated with designing effective multi-agent systems for ITS. We intentionally exclude discussions on the technical and significant challenges related to computational load and the limitations of AI models.
\subsubsection{Optimized Task Coordination and Orchestration}
The coordination of multiple agents poses significant challenges, particularly in optimizing both the domain workflow and computing load. These challenges arise from the need to balance the diverse tasks each agent performs while ensuring efficient use of computational resources. Conflicts may occur when agents compete for shared resources or when task dependencies create bottlenecks in the workflow. A potential solution to these challenges lies in applying game theory, which can model the interactions between agents as strategic games. By employing concepts like Nash equilibrium, agents can be guided to cooperate or compete in ways that optimize overall system performance, ensuring efficient task allocation and resource management \citep{wang2011integrating, lownes2011many}.

\subsubsection{Data sovereignty}
Data sovereignty concerns, including data ownership, colonialism, and residency, can impede the development of effective multi-agent systems for intelligent transportation by restricting access to crucial data. Data ownership laws can limit the sharing of information across borders, while fears of data colonialism may lead regions to withhold data to protect their interests. Additionally, data residency requirements, which mandate that data remain within specific geographic boundaries, can complicate the retrieval and processing of information across distributed systems. These challenges reduce the multi-agent system's ability to access comprehensive and timely data, ultimately limiting its effectiveness in optimizing transportation solutions \citep{xu2024leveraging}.

\subsubsection{AI accountability}
AI accountability is a critical challenge in building multi-agent systems for ITS. Ensuring that decisions made by autonomous agents can be traced, and responsibility attributed, is essential for addressing errors, biases, and unintended outcomes. Without clear accountability mechanisms, the reliability, safety, and ethical operation of ITS could be compromised, making it crucial to implement robust accountability measures to build trust and ensure system transparency.

\section{Concluding Remarks}
\label{sec:RemarksVision}
This paper has explored the transformative potential of integrating LLMs and RAG into multi-agent systems for ITS. Our proposed framework demonstrates how these technologies can enhance scalability, accessibility, and human-centric management in urban mobility. LLMs, combined with RAG and task-specific agents, offer new capabilities for autonomous task management, natural language interaction, and efficient coordination, paving the way for advanced smart city applications.

However, challenges such as task coordination, data sovereignty, and AI accountability must be addressed to fully realize these benefits. Overcoming these hurdles will be critical for leveraging AI-powered multi-agent systems to create safer, more efficient, and sustainable urban transportation systems. Future research should focus on refining these technologies and expanding their applications across broader smart city contexts.

\bibliographystyle{elsarticle-harv}
\bibliography{bib_file}

\end{document}